\documentclass[sn-chicago,iicol]{sn-jnl}

\usepackage{multirow}
\usepackage[table]{xcolor} 
\definecolor{cb}{rgb}{0.,0.,0.}

\def\*#1{\boldsymbol{#1}}

\jyear{2023}%

\theoremstyle{thmstyleone}%
\theoremstyle{thmstyletwo}%

\theoremstyle{thmstylethree}%

\begin{document}

\title[Article Title]{How Does Fine-Tuning Impact Out-of-Distribution Detection for Vision-Language Models?}


\author[1]{\fnm{Yifei} \sur{Ming}}\email{ming5@wisc.edu}

\author[1]{\fnm{Yixuan} \sur{Li}}\email{sharonli@cs.wisc.edu}

\affil[1]{\orgdiv{Department of Computer Sciences}, \orgname{University of Wisconsin-Madison}}

\abstract{ 
Recent large vision-language models such as CLIP have shown remarkable out-of-distribution (OOD)  detection and generalization performance. However, their zero-shot in-distribution (ID) accuracy is often limited for downstream datasets.
Recent CLIP-based fine-tuning methods such as prompt learning have demonstrated significant improvements in ID classification and OOD generalization where OOD labels are available. Nonetheless, it remains unclear whether the model is reliable to semantic shifts without OOD labels. In this paper, we aim to bridge the gap and present a comprehensive study to understand how fine-tuning impact OOD detection for few-shot downstream tasks. By framing OOD detection as multi-modal concept matching, we establish a connection between fine-tuning methods and various OOD scores. Our results suggest that a proper choice of OOD scores is essential for CLIP-based fine-tuning. In particular, the maximum concept matching (MCM) score provides a promising solution consistently. We also show that prompt learning demonstrates the state-of-the-art OOD detection performance over the zero-shot counterpart.
} 


\maketitle

\section{Introduction}\label{intro}
Machine learning (ML) is undergoing a paradigm shift with the rise of models that are trained on massive data and are adaptable to a wide range of downstream tasks. Popular pre-trained large vision-language models~\citep{radford2021learning,jia2021scaling,yao2021filip,li2021supervision} demonstrate remarkable performance, and allow researchers without extensive computation power to benefit from these models. It is now the common practice of the ML community to adopt pre-trained models for transfer learning on downstream tasks rather than learning from scratch. Despite the promise, the safety risks of these large pre-trained models can be potentially inherited by all the fine-tuned models. Without appropriately understanding the safety risks, development on top of pre-trained models can exacerbate and propagate safety concerns writ large, causing profound impacts on society.

In response to these urgent challenges, the overall objective of this paper is to systematically understand the out-of-distribution risks of learning with  pre-trained vision-language models. This paper seeks to address the research question that arises in building responsible and ethical AI models: 
\emph{How does fine-tuning influence out-of-distribution (OOD) detection for large vision-language models?}
Detecting OOD samples is crucial for machine learning models deployed in the open world, where samples from unseen classes naturally emerge, and failure to detect them can have severe consequences. 
Despite
increasing attention~\citep{yang2021generalized}, OOD detection research for large vision-language models has been scant. Among the most recent works, \cite{ming2022delving} investigated training-free OOD detection based on the pre-trained CLIP model. However, the impact of fine-tuning on OOD detection has been unexplored in the vision-language literature.

In this paper, we bridge the gap by investigating how fine-tuning large vision-language models affects OOD detection. 
Parameter-efficient fine-tuning methods have been popularized in recent years. In particular, prompt learning ~\citep{zhou2022cocoop, zhou2022coop} optimizes learnable word embeddings of the prompts, while adaptors directly optimize the internal feature representations~\citep{gao2021clip,zhang2022tip}. Both methods are parameter-efficient as image and text encoders are frozen during fine-tuning, and have shown significant improvement for few-shot in-distribution (ID) classification. Complementary to existing research, we focus on OOD detection for fine-tuned models using multi-modal concept matching. \textcolor{cb}{At the core of the concept matching} framework, we use the few-shot ID training set and textual descriptions of the labels to derive a set of visual and textual features that represent the typical features for each ID class. We can measure OOD uncertainty based on the distance between the input feature and the nearest ID prototype.
 
Based on \textcolor{cb}{the concept matching} framework, we then present a comprehensive and systematic study to explore how different parameter-efficient fine-tuning methods impact OOD detection performance, and contribute unexplored findings to the community. We disentangle various aspects such as adaptation methods and OOD scoring functions. 
Interestingly, we observe that parameter-efficient fine-tuning can significantly improve OOD reliability compared to zero-shot CLIP models. In particular, prompt learning methods exhibit very competitive performance \textcolor{cb}{when coupled with the maximum concept matching (MCM) score~\citep{ming2022delving}}. We term the overall scheme as \textbf{PEFT-MCM}, where PEFT indicates that the MCM score is applied to a model after parameter-efficient fine-tuning.
 
 Furthermore, we delve deeper into prompt learning and analyze how 
 the pre-trained features are modified during fine-tuning, and how it impacts OOD detection as a consequence. We study the impact of shots, architectures, and explore the effects of prompt learning on various downstream tasks, including the challenging ImageNet-1k (ID) benchmark. Our results demonstrate that prompt learning perturbs the pre-trained feature space that benefits both ID and OOD performance. More encouragingly, the trend holds consistently across different settings, highlighting its potential for reliable fine-tuning in vision-language modeling.
 
 We summarize the contributions of this work as follows:
 \begin{itemize}
     \item We provide a timely and systematic study on how CLIP-based fine-tuning influences OOD detection in the few-shot setting. Our study disentangles various factors, including adaptation methods and OOD scoring functions.
     \item We present novel evidence that parameter-efficient fine-tuning does not deteriorate pre-trained features. Instead, they can improve both ID and OOD performance with a proper OOD scoring function, \textcolor{cb}{especially the MCM score} (\emph{i.e.,} PEFT-MCM). We show that prompt learning consistently demonstrates the state-of-the-art OOD detection performance over the zero-shot counterpart.
     \item  We provide an in-depth analysis of prompt learning's impact on the feature space for OOD detection and conduct comprehensive ablations across datasets, architectures, and the number of shots with various OOD detection scores. 
 \end{itemize}

\begin{figure*}[htb]
  \centering
    \includegraphics[width=1.0\linewidth]{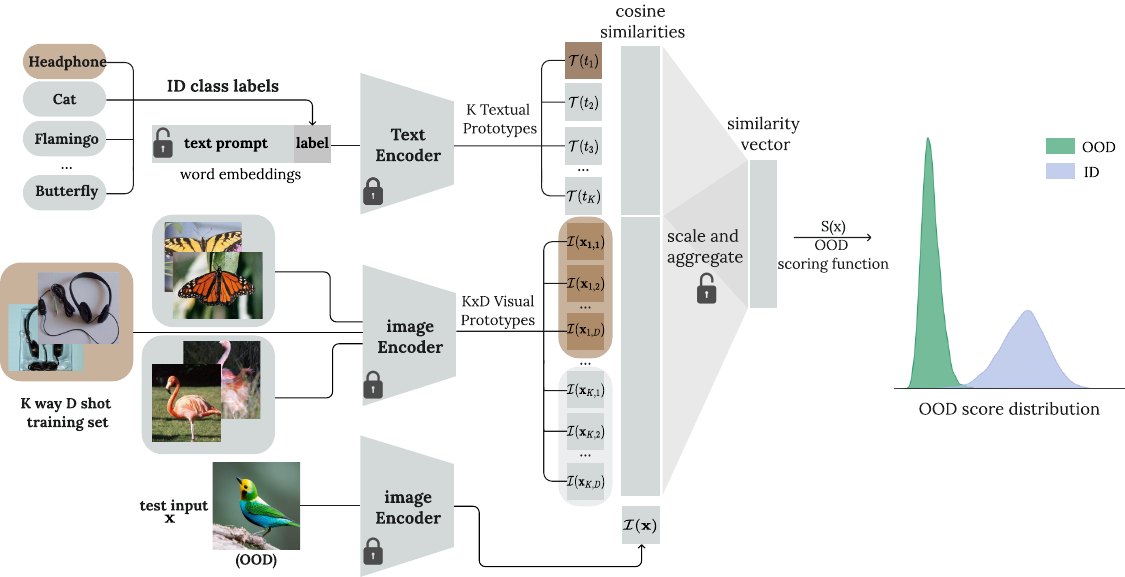}
    
\caption{ A unified pipeline for OOD detection with parameter-efficient fine-tuning of CLIP models on few-shot datasets. Given ID text labels $\mathcal{Y}_\text{in}$ and a few-shot training set, we view the textual and visual embeddings of ID classes as concept prototypes in the feature space. The OOD uncertainty of an input image can be characterized by the distance from its visual feature to the closest ID prototype from both modalities. See Section~\ref{method} for details.}
\vspace{-3mm}
\label{fig:teaser}
\end{figure*}

\section{Preliminaries}\label{prelim}
\noindent\textbf{Contrastive vision-language models.}
Recent large vision-language models have shown great potential for various computer vision tasks. In this paper, we focus on CLIP-like models~\citep{radford2021learning,yao2021filip}, which adopt a dual-stream architecture with one text encoder $f: t \rightarrow \mathbb{R}^d$ and one image encoder $g: \*x \rightarrow \mathbb{R}^d$. CLIP is pre-trained on a massive web-scale image-caption dataset with a multi-modal contrastive loss that promotes the alignment of features from different modalities. CLIP learns transferable feature representations and demonstrates promising zero-shot generalization performance~\citep{fort2021exploring}. Despite the promise, existing vision-language models perform zero-shot
classification in a \emph{closed-world} setting. That is, it will match an input into a fixed set of categories,
even if it is irrelevant. For example, a bird in Figure~\ref{fig:teaser} can  be blindly predicted as one of the in-distribution classes $\mathcal{Y}_\text{in}=$\{headphone, cat, flamingo, butterfly\}. This motivates the importance of OOD detection for vision-language models.

\vspace{0.2cm}
\noindent\textbf{OOD detection for vision-language models.} In the open-world setting, the goal of OOD detection is to detect samples that do not belong to  ID classes $\mathcal{Y}_\text{in}$. Here ID classes are
defined w.r.t. the classification task of interest, instead of the classes used in pre-training. Accordingly,
OOD is defined w.r.t. the ID classes, not the data distribution during pre-training. ~\cite{ming2022delving} explore the zero-shot OOD detection for the pre-trained CLIP model, without adapting to the ID dataset. Instead, we focus on the setting where CLIP models are fine-tuned on a few-shot dataset $\mathcal{D}_{\text{in}}$, and hence are better adapted to the downstream ID task. We evaluate the fine-tuned CLIP model on a combination of ID and OOD datasets $\mathcal{D}_{\text{in}} \cup \mathcal{D}_{\text{out}}$, where $\mathcal{D}_{\text{out}} = \{\*x_i, y_i^{\text{out}}\}_{i=1}^m$ contains inputs with semantically different categories $y^{\text{out}}\notin \mathcal{Y}_\text{in}$.  Formally, given an input $\*x$, 
OOD detection can be formulated as:
 \begin{align*}
\label{eq:threshold}
	G(\*x; f, g) =\begin{cases} 
      1 & S(\*x;f, g)\ge \lambda \\
      -1 & S(\*x;f, g) < \lambda 
  \end{cases},
\end{align*}
where $S(\cdot)$ is a scoring function that measures OOD uncertainty. In practice, $\lambda$ is chosen so that a high fraction of ID data (\textit{e.g.}, 95\%) is above the threshold.

\vspace{0.2cm}
\noindent\textbf{Parameter-efficient fine-tuning.} To improve the performance on downstream tasks, parameter-efficient approaches are proposed to fine-tune CLIP on datasets of interest. Prompt learning and adaptor tuning have recently gained popularity and demonstrated improved results over zero-shot settings. In particular, prompt learning optimizes the word embeddings of the prompts, while adaptors directly optimize the internal feature representations. Both methods are parameter-efficient as image and text encoders are frozen during fine-tuning. \textcolor{cb}{In what follows, we introduce prompt-based and adaptor-based methods respectively.}

For a downstream dataset with $K$ in-distribution classes $\mathcal{Y}_\text{in}=\{y_1, y_2,...,y_K\}$, prompt learning method such as \texttt{CoOp}~\citep{zhou2022coop} 
introduces $M$ learnable context vectors $v_i\in \mathbb{R}^e$ to replace hand-engineered text prompts such as
``\texttt{this is a photo of}'', where $e$ is the dimension of word embeddings. For each class $y_k$, we obtain its contextualized representation $t_k = [v_1, v_2, \cdots, v_M, w_k]$ by concatenating the context vectors and the word embedding $w_k\in \mathbb{R}^e$ of the label (upper left, Figure~\ref{fig:teaser}). To avoid overfitting and improve generalization performance, \texttt{CoCoOp}~\citep{zhou2022cocoop} further introduces instance-conditional prompts via a meta-network which produces a meta token $m(\*x)$ given the visual feature of the input $\*x$. The meta token is added to each context token $v_i(\*x) = v_i + m(\*x)$ for $i \in \{1,2,\cdots, M\}$. Therefore, the prompt for class $k$ is conditioned on each input: $t_k(\*x) = [v_1{(\*x)}, v_2{(\*x)}, \cdots, v_M{(\*x)}, w_k]$.  
 To learn the context vectors, the cross-entropy loss is used in fine-tuning:
\begin{equation}
p(y_k\mid \boldsymbol{x})=\frac{\exp \left(s_k(\*x) / \tau\right)}{\sum_{i=1}^K \exp \left( s_i(\*x)/ \tau\right)},
\end{equation}
where $s_k(\*x) = \frac{g(\*x) \cdot f(t_k)}{\lVert g(\*x)\rVert \cdot \lVert f(t_k) \rVert}$ is the cosine similarity of input $\*x$ with the $k$-th label, and $\tau$ is the temperature.

Alternatively, adaptor-based methods directly optimize the feature representations $g(\*x)$ instead of learning context vectors. Specifically, given a $K$-way-$D$-shot ID training set (consisting of $K$ classes with $D$ examples per class), ~\cite{zhang2022tip} propose a training-free adaptation method \texttt{TipAdaptor} which extracts all the visual features $W_g = [g(\*x_{1,1}),g(\*x_{1,2}),\cdots, g(\*x_{K,D})]\in \mathbb{R}^{KD \times d}$ from the few-shot training dataset. For each input $\*x$, we can obtain $K\times D$ cosine similarities 
$s_{k,d}(\*x) = \frac{g(\*x) \cdot g(\*x_{k,d})} {\lVert g(\*x)\rVert \cdot \lVert g(\*x_{k,d}) \rVert}$. The cosine similarities are scaled by an exponential function $\tilde{s}: s \mapsto \exp(-\beta + \beta s)$  with a hyperparameter $\beta$ that modulates the sharpness. Therefore,  we can obtain an average similarity vector for each class based on visual features, $\tilde{s}_k(\*x) = { 1\over D}\sum_{d=1}^D \tilde{s}_{k,d}(\*x) $. The final similarity for class $k$ is a weighted sum of similarities from the two modalities $\alpha \tilde{s}_k(\*x)  + s_k(\*x) $. To achieve better few-shot ID performance, \cite{zhang2022tip} set visual features $W_g$ as learnable parameters and denote the method as \texttt{TipAdaptorF}, where F stands for fine-tuning.
Despite the stronger downstream classification performance, it remains unknown if fine-tuning leads to more reliable OOD detection at test time.
We aim to provide a comprehensive understanding in this paper.

\section{Method}\label{method}
\subsection{OOD detection with fine-tuning}
We investigate OOD detection with parameter-efficient fine-tuning on downstream tasks. We present a unified framework in Figure~\ref{fig:teaser}, where the learnable part of the CLIP model is marked with an ``unlock'' icon while the frozen part is marked with a ``lock'' icon. For prompt learning methods such as \texttt{CoOp} and \texttt{CoCoOp}, the cosine similarity of the input feature with the $k$-th class $s_k(\*x) = \frac{g(\*x) \cdot f(t_k)}{\lVert g(\*x)\rVert \cdot \lVert f(t_k) \rVert}$ is derived based on the adapted textual feature vector $t_k$.
Alternatively, adaptor-based methods such as \texttt{TipAdaptor} and \texttt{TipAdaptorF} first scale the cosine similarities of visual prototypes and perform a weighted sum with the similarities of textual prototypes. Therefore, we can view \texttt{TipAdaptor} as an ensemble method that utilizes multi-modal prototypes. 

To summarize, for each adaptation algorithm $\mathcal{A}$, OOD detection can be performed by:
 \begin{align*}
\label{eq:threshold}
	G_\mathcal{A}(\*x; f, g) =\begin{cases} 
      \text{ID} & S(\*x;f, g)\ge \lambda \\
      \text{OOD} & S(\*x;f, g) < \lambda 
  \end{cases},
\end{align*}
where $\mathcal{A}$ can be instantiated by an adaptation method such as \texttt{CoOp}, \texttt{CoCoOp}, \texttt{TipAdaptor}, or \texttt{TipAdaptorF}. Therefore, the OOD detector $G_\mathcal{A}(\cdot)$ can be viewed as a
``safeguard'' for the classification model.  \textcolor{cb}{Next, we introduce various OOD score functions $S(\*x;f, g)$ assuming $G_\mathcal{A}(\*x; f, g)$ is defined implicitly as each score function corresponds to an OOD detector $G$. }

\subsection{OOD score for vision-language models}
\label{sec:score}
Recently,~\cite{ming2022delving} propose a conceptual framework of CLIP-based OOD detection via concept matching, where the textual feature $f(t_k)$ is viewed as the concept prototype for ID class $k\in \{1,2,...,K\}$. OOD uncertainty is then characterized by the distance from the visual feature of the input to the closest ID textual prototype. That is, images closer to one of the ID prototypes are more likely to be ID and vice versa. ~\cite{ming2022delving} suggest that softmax scaling with a proper temperature $\tau$ provably leads to state-of-the-art performance under the zero-shot (training-free) setting. Specifically, the maximum concept matching (MCM) score is defined as:
\begin{equation}
    S_{\text{MCM}}(\*x) = \max_{k \in [K]} \frac{e^{s_k(\*x)/\tau}}{\sum_{j=1}^K e^{s_j(\*x)/\tau}},
\end{equation}
where the temperature $\tau$ needs to be tuned on the downstream dataset. \textcolor{cb}{As a special case of MCM, we use MSP to denote the MCM score when the temperature $\tau_d$ is set as default for CLIP models at inference time (\emph{e.g.,} 100 for CLIP-B/16}).

Additionally, we consider a simpler scoring function based on the maximum similarity (MS) among ID prototypes before applying softmax scaling: \begin{equation}S_{\text{MS}}(\*x) = \max_{k\in[K]} s_k(\*x),\end{equation}
which does not require any hyperparameter tuning. We show in Section~\ref{exp} that the MS score demonstrates strong OOD detection performance with fine-tuning, especially for fine-grained ID datasets. 
We now proceed to experiments where we investigate the impact of fine-tuning on real-world tasks.

\section{Experiments}\label{exp}

\begin{table*}[htb]
\caption{OOD detection performance based on $S_{\text{MS}}$ score (w.o. softmax scaling). When ID datasets contain finer-grained categories semantically different from OOD categories, the pre-trained CLIP model demonstrates nearly perfect OOD detection performance. More encouragingly, after adapting the model to downstream datasets, OOD detection performance remains competitive.}
\label{tab:fine-grain}
\centering
\resizebox{0.95\textwidth}{!}{
\begin{tabular}{llcccccccccc}
\toprule
\multirow{2}{*}{\textbf{ID Dataset}}           & \multirow{2}{*}{\textbf{Method}} & \multicolumn{2}{c}{\textbf{SUN}} & \multicolumn{2}{c}{\textbf{Places}} & \multicolumn{2}{c}{\textbf{Textures}} & \multicolumn{2}{c}{\textbf{iNaturalist}} & \multicolumn{2}{c}{\textbf{Average}} \\
\cmidrule(lr){3-4} \cmidrule(lr){5-6} \cmidrule(lr){7-8} \cmidrule(lr){9-10} \cmidrule(lr){11-12}
                              &                         & \textbf{FPR95$\downarrow$}         & \textbf{AUROC$\uparrow$}     & \textbf{FPR95$\downarrow$}         & \textbf{AUROC$\uparrow$}       & \textbf{FPR95$\downarrow$}         & \textbf{AUROC$\uparrow$}         & \textbf{FPR95$\downarrow$}         & \textbf{AUROC$\uparrow$}          & \textbf{FPR95$\downarrow$}         & \textbf{AUROC$\uparrow$}      \\
                              \midrule
                              &\multicolumn{11}{c}{\textbf{Training not required}}   \\
\multirow{7}{*}{Food-101}     
                             & ZOCLIP                  & 0.04       & 99.92      & 0.12         & 99.93       & 4.63         & 98.29         & 0.15           & 99.87          & 1.24       & 99.50      \\
& TipAdaptor                 & 0.00       & 99.94      & 0.04         & 99.95       & 2.87         & 98.85         & 0.06           & 99.90          & 0.74       & 99.66      \\
&\multicolumn{11}{c}{\textbf{Requires training}}          \\
                              & TipAdaptorF                & 0.00       & 99.94      & 0.03         & 99.95       & 3.16         & 98.77         & 0.05           & 99.91          & 0.81       & 99.64      \\
                              & CoOp                    & 0.01       & 99.97      & 0.00         & 99.98       & 1.45         & 99.68         & 0.00           & 99.97          & 0.36       & 99.90      \\
                              & CoCoOp                  & 0.00       & 99.98      & 0.00         & 99.98       & 1.97         & 99.51         & 0.01           & 99.97          & 0.49       & 99.86      \\
                               \midrule
                               &\multicolumn{11}{c}{\textbf{Training not required}}   \\
\multirow{7}{*}{Oxford-Pets}                     & ZOCLIP                  & 0.03       & 99.99      & 0.14         & 99.96       & 0.12         & 99.95         & 0.00           & 100.00         & 0.07       & 99.97      \\
&TipAdaptor                 & 0.01       & 100.00     & 0.07         & 99.98       & 0.07         & 99.99         & 0.00           & 100.00         & 0.04       & 99.99      \\
&\multicolumn{11}{c}{\textbf{Requires training}}          \\
                              & TipAdaptorF                & 0.02       & 100.00     & 0.07         & 99.98       & 0.09         & 99.98         & 0.00           & 100.00         & 0.04       & 99.99      \\
                              & CoOp                    & 0.02       & 100.00     & 0.18         & 99.97       & 0.25         & 99.92         & 0.00           & 100.00         & 0.11       & 99.97      \\
                              & CoCoOp                  & 0.03       & 99.99      & 0.19         & 99.96       & 0.11         & 99.96         & 0.00           & 100.00         & 0.08       & 99.98      \\
                              \midrule
                              &\multicolumn{11}{c}{\textbf{Training not required}}   \\
\multirow{7}{*}{Stanford-Cars}                  & ZOCLIP                  & 0.02       & 99.99      & 0.24         & 99.94       & 0.00         & 100.00        & 0.00           & 100.00         & 0.07       & 99.98      \\
& TipAdaptor                 & 0.01       & 100.00     & 0.08         & 99.98       & 0.00         & 100.00        & 0.00           & 100.00         & 0.02       & 100.00     \\
             &\multicolumn{11}{c}{\textbf{Requires training}}          \\
                              & TipAdaptorF                & 0.01       & 100.00     & 0.06         & 99.98       & 0.00         & 100.00        & 0.00           & 100.00         & 0.02       & 100.00     \\
                              & CoOp                    & 0.01       & 100.00     & 0.07         & 99.97       & 0.00         & 100.00        & 0.00           & 100.00         & 0.02       & 99.99      \\
                              & CoCoOp                  & 0.01       & 100.00     & 0.07         & 99.97       & 0.00         & 100.00        & 0.00           & 100.00         & 0.02       & 99.99      \\
             \midrule
              &\multicolumn{11}{c}{\textbf{Training not required}}   \\
                                       \multirow{7}{*}{Caltech-101}                                & ZOCLIP                  & 32.03      & 94.06      & 33.01        & 93.39       & 54.66        & 89.29         & 32.14          & 94.30          & 37.96      & 92.76      \\
                              & TipAdaptor                 & 9.69       & 98.07      & 11.25        & 97.84       & 20.90        & 96.68         & 13.62          & 97.72          & 13.86      & 97.58      \\
                               &\multicolumn{11}{c}{\textbf{Requires training}}          \\
                              & TipAdaptorF                & 10.20      & 97.76      & 11.60        & 97.42       & 23.32        & 95.54         & 14.01          & 97.36          & 14.78      & 97.02      \\
                              & CoOp                    & 5.53       & 98.56      & 9.88         & 97.50       & 13.10        & 97.10         & 4.89           & 98.76          & 8.35       & 97.98      \\
                              & CoCoOp                  & 2.86       & 99.19      & 6.42         & 98.37       & 8.81         & 98.09         & 5.68           & 98.68          & \textbf{5.94}       & \textbf{98.58}      \\
                               
                              \bottomrule
                              
\end{tabular}
}
\end{table*}

\subsection{Setup}\label{exp_setup}

\vspace{0.2cm}
\noindent\textbf{Datasets.} Following \cite{ming2022delving},  
we consider a wide range of real-world ID datasets with various semantics and number of classes: Caltech-101~\citep{bossard14}, Stanford-Cars~\citep{KrauseStarkDengFei-Fei_3DRR2013}, Food-101~\citep{bossard14},  Oxford-Pets~\citep{parkhi12a} and ImageNet-1k~\citep{deng2009imagenet}.
For each ID dataset, we follow~\cite{zhou2022cocoop} and construct the training set with $D$ random samples per class, while the original test set is used for testing. We use $D=16$ by default and study the impact of shots as ablations in Section~\ref{sec:delve}.
For OOD test datasets, we use the same ones in~\cite{huang2021mos}, including subsets of iNaturalist~\citep{van2018inaturalist}, \textsc{Sun}~\citep{xiao2010sun}, \textsc{Places}~\citep{zhou2017places}, and \textsc{Texture}~\citep{cimpoi2014describing}. 
For each OOD dataset, the categories do not overlap with the ID dataset. \textcolor{cb}{For ImageNet-1k as ID, we also consider two additional OOD datasets ImageNet-O~\citep{hendrycks2021nae} and OpenImage-O~\citep{wang2022vim}.}

\vspace{0.2cm}
\noindent\textbf{Models and training details.} For pre-trained models, we use CLIP-B/16 as the default backbone for main experiments, which uses ViT-B/16~\citep{dosovitskiy2021an} as the image encoder. The impact of backbones is included in the ablation studies. 
We use \texttt{ZOCLIP} to denote pre-trained CLIP without fine-tuning. For each method, we closely follow the original implementations. Specifically, for \texttt{CoOp} and \texttt{CoCoOp}, the context length is set to 4, and the context vectors are initialized using the pre-trained word embeddings of ``a photo of a''. \texttt{CoCoOp} is 
 trained with a batch size of 1 for 10 epochs using SGD, while \texttt{CoOp} is trained for 100 epochs with a batch size of 32. \texttt{TipAdapterF} is trained with a batch size 256 using AdamW~\citep{loshchilov2018decoupled} for 20 epochs. Cosine scheduling is used for all methods and the data preprocessing protocol consists of random re-sizing, cropping, and random horizontal flip.

\vspace{0.2cm}
\noindent\textbf{Evaluation metrics.} We consider the following evaluation metrics: (1) the false positive rate (\text{FPR}95) of OOD samples when the true positive rate of in-distribution samples is at 95\%, (2) the area under the receiver operating characteristic curve (AUROC), and (3) ID classification accuracy (ID ACC). 

\subsection{Main results and discussions}\label{main_results}
\textcolor{cb}{In this section, we first present novel evidence that parameter-efficient fine-tuning generally improves OOD performance over the zero-shot counterpart with a simple OOD scoring function. Next, we investigate the effects of various OOD scoring functions in the parameter-efficient fine-tuning setting. In particular, we will show that the MCM score consistently demonstrates the most promising performance compared to alternative OOD scores when coupled with prompt learning.}

\vspace{0.2cm}
\noindent\textbf{How does parameter-efficient fine-tuning impact OOD detection?}  We evaluate the OOD detection performance on various ID datasets. The results are summarized in Table~\ref{tab:fine-grain}.
We show that adapted CLIP models demonstrate nearly perfect OOD detection performance for ID datasets with fine-grained categories such as Stanford-Cars and Oxford-Pets. 
Moreover, when the ID dataset contains a diverse collection of categories such as Caltech-101\footnote{Similar trends also hold for ImageNet-1k as ID.}, 
parameter-efficient fine-tuning still significantly improves the OOD detection performance on average compared to \texttt{ZOCLIP}. In particular, \texttt{CoCoOp} yields the best performance among other adaptation methods on Caltech-101 (ID). It achieves an average FPR95 of 5.94\% using $S_{\text{MS}}$, improving by $32.02\%$ over \texttt{ZOCLIP}. 
While prior works suggest that parameter-efficient fine-tuning methods improve ID accuracy on few-shot datasets, our results complement their findings and show that fine-tuning also improves the OOD detection performance with proper OOD scoring functions.

\begin{table*}[!bht]
\caption{OOD detection performance with $S_{\text{MS}}$ and $S_{\text{MCM}}$ score when the ID dataset contains diverse categories. Prompt learning methods display clear advantages over zero-shot models. The results are based on Caltech-101 (ID). 
}
\label{tab:comparison}
\centering
\resizebox{0.95\textwidth}{!}{
\begin{tabular}{clcccccccccc}
\toprule
\multirow{2}{*}{\textbf{OOD Score}}           & \multirow{2}{*}{\textbf{Method}} & \multicolumn{2}{c}{\textbf{SUN}} & \multicolumn{2}{c}{\textbf{Places}} & \multicolumn{2}{c}{\textbf{Textures}} & \multicolumn{2}{c}{\textbf{iNaturalist}} & \multicolumn{2}{c}{\textbf{Average}} \\
\cmidrule(lr){3-4} \cmidrule(lr){5-6} \cmidrule(lr){7-8} \cmidrule(lr){9-10} \cmidrule(lr){11-12}
                              &                         & \textbf{FPR95$\downarrow$}         & \textbf{AUROC$\uparrow$}     & \textbf{FPR95$\downarrow$}         & \textbf{AUROC$\uparrow$}       & \textbf{FPR95$\downarrow$}         & \textbf{AUROC$\uparrow$}         & \textbf{FPR95$\downarrow$}         & \textbf{AUROC$\uparrow$}          & \textbf{FPR95$\downarrow$}         & \textbf{AUROC$\uparrow$}      \\
                              \midrule
                              \multirow{5}{*}{$S_{\text{MS}}$}                                & ZOCLIP                  & 32.03      & 94.06      & 33.01        & 93.39       & 54.66        & 89.29         & 32.14          & 94.30          & 37.96      & 92.76      \\
                              & TipAdaptor                 & 9.69       & 98.07      & 11.25        & 97.84       & 20.90        & 96.68         & 13.62          & 97.72          & 13.86      & 97.58      \\
                              & TipAdaptorF                & 10.20      & 97.76      & 11.60        & 97.42       & 23.32        & 95.54         & 14.01          & 97.36          & 14.78      & 97.02      \\
                              & CoOp                    & 5.53       & 98.56      & 9.88         & 97.50       & 13.10        & 97.10         & 4.89           & 98.76          & 8.35       & 97.98      \\
                              & CoCoOp                  & 2.86       & 99.19      & 6.42         & 98.37       & 8.81         & 98.09         & 5.68           & 98.68          & \textbf{5.94}       & \textbf{98.58}      \\
                               \midrule
                              \multirow{5}{*}{$S_{\text{MCM}}$}                & ZOCLIP                 &14.83&	97.20 &20.45&	96.00 &	14.98&	97.35	&10.84&	97.76&	15.28&	97.08    \\
& TipAdaptor    &  5.12	&98.83    &8.05&	98.34&	4.65&	99.05&	6.94&	98.77&	6.19	&\textbf{98.75}   \\
                              & TipAdaptorF &4.83	&98.79 & 8.09&	98.07	&6.41	&98.11	&4.94	&98.98&	6.07&	98.49                  \\
                
                              & CoOp    &     3.62& 	99.01	& 8.15& 	97.89	& 6.29& 	98.62& 	7.57	& 98.35	& 6.41	& 98.47                   \\
                              & CoCoOp   &   4.26&	98.94	&6.76&	98.00&	4.33	&98.88&	4.71&	98.68&	\textbf{5.02}&	98.62                 \\
                              \bottomrule
                              
\end{tabular}
}
\end{table*}

\begin{figure}[!htb]
  \centering
    \includegraphics[width=0.95\linewidth]{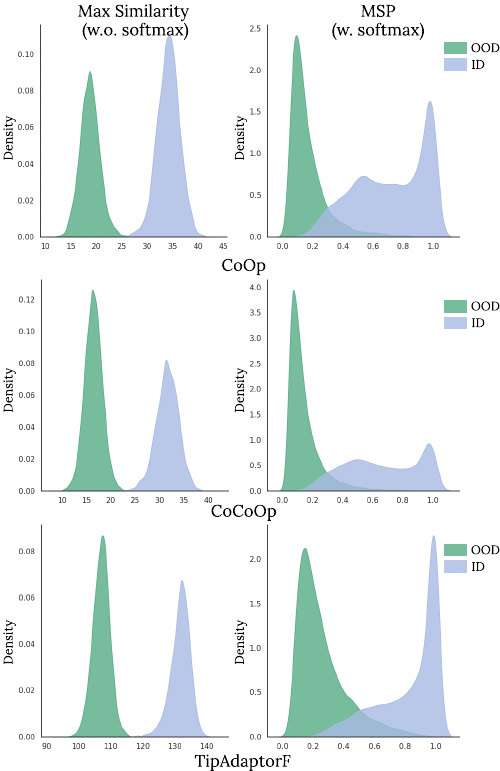}
\caption{The impact of softmax scaling. We use Stanford-Cars (ID) vs. SUN (OOD) for illustration. Applying softmax scaling significantly decreases ID-OOD separability for \texttt{CoOp} (top row), \texttt{CoCoOp} (second row), and \texttt{TipAdaptorF} (last row), resulting in worse OOD detection performance.}
\label{fig:softmax}
\end{figure}

\begin{table*}[htb]
\caption{OOD detection performance based on $S_{\text{MCM}}$ score.}
\label{tab:fine-grain_mcm}
\centering
\resizebox{0.95\textwidth}{!}{
\begin{tabular}{llcccccccccc}
\toprule
\multirow{2}{*}{\textbf{ID Dataset}}           & \multirow{2}{*}{\textbf{Method}} & \multicolumn{2}{c}{\textbf{SUN}} & \multicolumn{2}{c}{\textbf{Places}} & \multicolumn{2}{c}{\textbf{Textures}} & \multicolumn{2}{c}{\textbf{iNaturalist}} & \multicolumn{2}{c}{\textbf{Average}} \\
\cmidrule(lr){3-4} \cmidrule(lr){5-6} \cmidrule(lr){7-8} \cmidrule(lr){9-10} \cmidrule(lr){11-12}
                              &                         & \textbf{FPR95$\downarrow$}         & \textbf{AUROC$\uparrow$}     & \textbf{FPR95$\downarrow$}         & \textbf{AUROC$\uparrow$}       & \textbf{FPR95$\downarrow$}         & \textbf{AUROC$\uparrow$}         & \textbf{FPR95$\downarrow$}         & \textbf{AUROC$\uparrow$}          & \textbf{FPR95$\downarrow$}         & \textbf{AUROC$\uparrow$}      \\                      
                              \midrule
\multirow{5}{*}{Food-101}      & ZOCLIP                  & 1.75       & 99.46      & 2.04         & 99.35       & 5.54          & 98.05        & 2.80           & 99.17          & 3.03       & 99.01      \\
                              & TipAdaptor              & 0.63       & 99.75      & 0.64         & 99.71       & 3.76          & 98.59        & 1.32           & 99.55          & \textbf{1.59}       & \textbf{99.40}      \\
                              & TipAdaptorF             & 1.77       & 99.57      & 1.57         & 99.53       & 4.43          & 98.34        & 1.85           & 99.40          & 2.40       & 99.21      \\
                              & CoOp                    & 2.00       & 99.46      & 1.60         & 99.47       & 5.85          & 98.39        & 1.37           & 99.54          & 2.71       & 99.22      \\
                              & CoCoOp                  & 1.06       & 99.69      & 1.01         & 99.63       & 4.17          & 98.42        & 1.40           & 99.53          & 1.91       & 99.32      \\
                              \midrule
\multirow{5}{*}{Oxford-Pets} & ZOCLIP                  & 1.18       & 99.73      & 3.37         & 99.28       & 1.37          & 99.73        & 6.17           & 98.84          & 3.02       & 99.40      \\
                              & TipAdaptor              & 0.05       & 99.97      & 0.62         & 99.87       & 0.17          & 99.96        & 0.11           & 99.87          & \textbf{0.24}       & \textbf{99.92}      \\
                              & TipAdaptorF             & 0.48       & 99.89      & 1.74         & 99.66       & 0.43          & 99.88        & 0.93           & 99.53          & 0.90       & 99.74      \\
                              & CoOp                    & 0.06       & 99.96      & 0.55         & 99.85       & 0.39          & 99.90        & 2.07           & 99.37          & 0.77       & 99.77      \\
                              & CoCoOp                  & 0.08       & 99.95      & 0.53         & 99.85       & 0.25          & 99.91        & 1.12           & 99.55          & 0.49      & 99.82      \\
                              \midrule
\multirow{5}{*}{Stanford-Cars} & ZOCLIP                  & 0.02       & 99.96      & 0.31         & 99.89       & 0.02          & 99.96        & 0.10           & 99.74          & 0.11       & 99.89      \\
                              & TipAdaptor              & 0.01       & 99.98      & 0.11         & 99.94       & 0.00          & 99.97        & 0.00           & 99.84          & \textbf{0.03}       & 99.93      \\
                              & TipAdaptorF             & 0.03       & 99.98      & 0.19         & 99.94       & 0.00          & 99.99        & 0.00           & 99.93          & 0.06       & \textbf{99.96}      \\
                              & CoOp                    & 0.01       & 99.98      & 0.17         & 99.93       & 0.00          & 99.98        & 0.02           & 99.84          & 0.05       & 99.93      \\
                              & CoCoOp                  & 0.02       & 99.98      & 0.15         & 99.93       & 0.00          & 99.97        & 0.00           & 99.87          & 0.04       & 99.94      \\
                               \midrule
                               \multirow{5}{*}{Caltech-101}                & ZOCLIP                 &14.83&	97.20 &20.45&	96.00 &	14.98&	97.35	&10.84&	97.76&	15.28&	97.08    \\
& TipAdaptor    &  5.12	&98.83    &8.05&	98.34&	4.65&	99.05&	6.94&	98.77&	6.19	&\textbf{98.75}   \\
                              & TipAdaptorF &4.83	&98.79 & 8.09&	98.07	&6.41	&98.11	&4.94	&98.98&	6.07&	98.49                  \\
                
                              & CoOp    &     3.62& 	99.01	& 8.15& 	97.89	& 6.29& 	98.62& 	7.57	& 98.35	& 6.41	& 98.47                   \\
                              & CoCoOp   &   4.26&	98.94	&6.76&	98.00&	4.33	&98.88&	4.71&	98.68&	\textbf{5.02}&	98.62                 \\
                              \bottomrule
                              
\end{tabular}
}
\end{table*}

\vspace{0.3cm}
\noindent\textbf{Effects of OOD scoring functions.} We investigate the effect of OOD scoring functions under fine-tuned vision-language models. In Table~\ref{tab:comparison}, we contrast the OOD detection performance using MCM~\citep{ming2022delving} vs. MS on Caltech-101 (ID). 
Our findings suggest that: (1) $S_{\text{MCM}}$ performs on par with $S_{\text{MS}}$ for fine-grained ID tasks across a wide range of adaptation methods (Table~\ref{tab:fine-grain_mcm}). (2) However, when ID contains diverse categories, utilizing  $S_{\text{MCM}}$ generally leads to better performance compared to using $S_{\text{MS}}$ for most adaptation methods (Table~\ref{tab:comparison}). (3) In particular, prompt learning methods such as \texttt{CoCoOp} demonstrate very competitive results with both OOD scores (an average FPR95 of $5.02$\%  with $S_{\text{MCM}}$ and $5.94$\% with $S_{\text{MS}}$ in Table~\ref{tab:comparison}).

\vspace{0.3cm}
\noindent\textbf{Effects of softmax scaling.} 
Previously, \citet{ming2022delving} observed that the commonly used maximum softmax score ($S_{\text{MSP}}$) is suboptimal for zero-shot OOD detection with vision-language models. 
We investigate whether MSP is competitive for OOD detection with fine-tuned models. 
To better illustrate the effects, we plot the score distributions for Stanford-Cars (ID) vs. SUN (OOD) in Figure~\ref{fig:softmax} when the model is fine-tuned with \texttt{CoOp}, \texttt{CoCoOp}, and \texttt{TipAdaptorF} respectively.
For each fine-tuning method, we can clearly see that the $S_{\text{MS}}$  leads to superior ID-OOD separability, while $S_{\text{MSP}}$  displays significant overlapping. Quantitatively, compared to $S_{\text{MSP}}$, the average FPR95 is significantly decreased with $S_{\text{MS}}$ (Table~\ref{tab:msp}). Our findings highlight that directly applying MSP is not competitive for fine-tuned vision-language models.

\subsection{Delving into parameter-efficient fine-tuning for OOD detection}  \label{sec:delve}
\vspace{0.2cm}

\begin{figure}[!htb]
  \centering
    \includegraphics[width=0.95\linewidth]{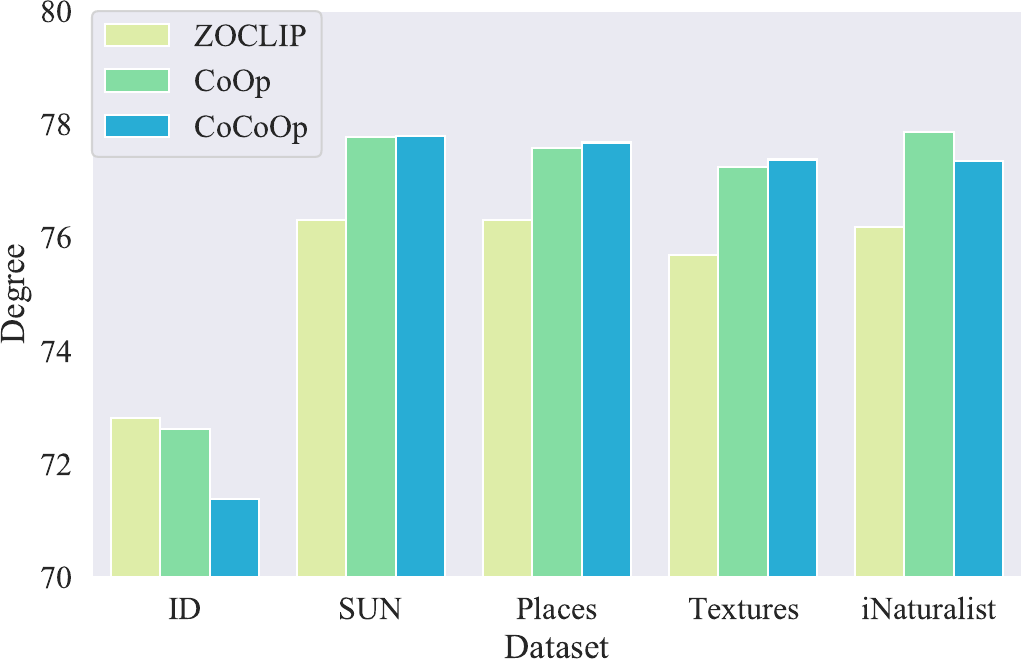}
\caption{Average $S_{\text{MS}}$ for ID (Caltech-101) and OOD test sets. Prompt learning methods decrease the angular distance for ID inputs while increasing the angular distance for OOD inputs to the nearest concept prototype, leading to better ID-OOD separability (Figure~\ref{fig:illustrate_sep}).}
\label{fig:id_ood_sep}
\end{figure}

\begin{figure}[!htb]
  \centering
    \includegraphics[width=0.96\linewidth]{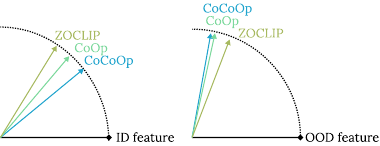}
\caption{Illustration of how prompt learning methods impact the hyperspherical features. Left: feature of an ID sample and its nearest ID prototype; Right: feature of an OOD sample and its nearest  ID prototype. }
\label{fig:illustrate_sep}
\end{figure}

\begin{figure}[!hbt]
  \centering
    \includegraphics[width=0.95\linewidth]{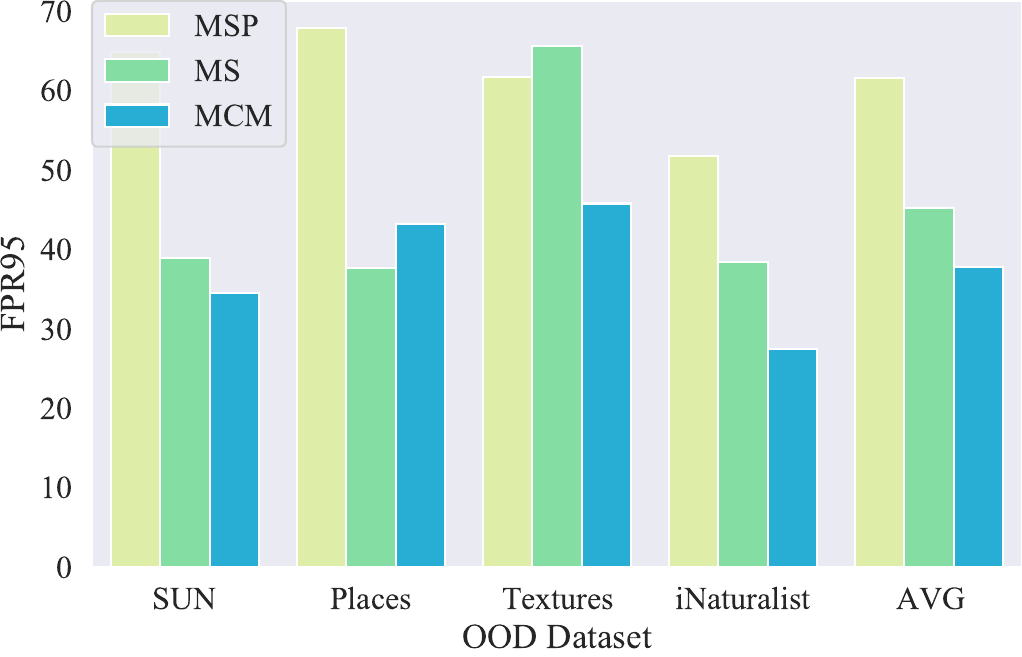}
\caption{OOD detection performance \textcolor{cb}{(FPR95)} on ImageNet-1k (ID). Using $S_{\text{MCM}}$ score leads to significant improvement over $S_{\text{MSP}}$.}
\label{fig:imagenet_fpr}
\end{figure}

\begin{figure}[!hbt]
  \centering
    \includegraphics[width=0.95\linewidth]{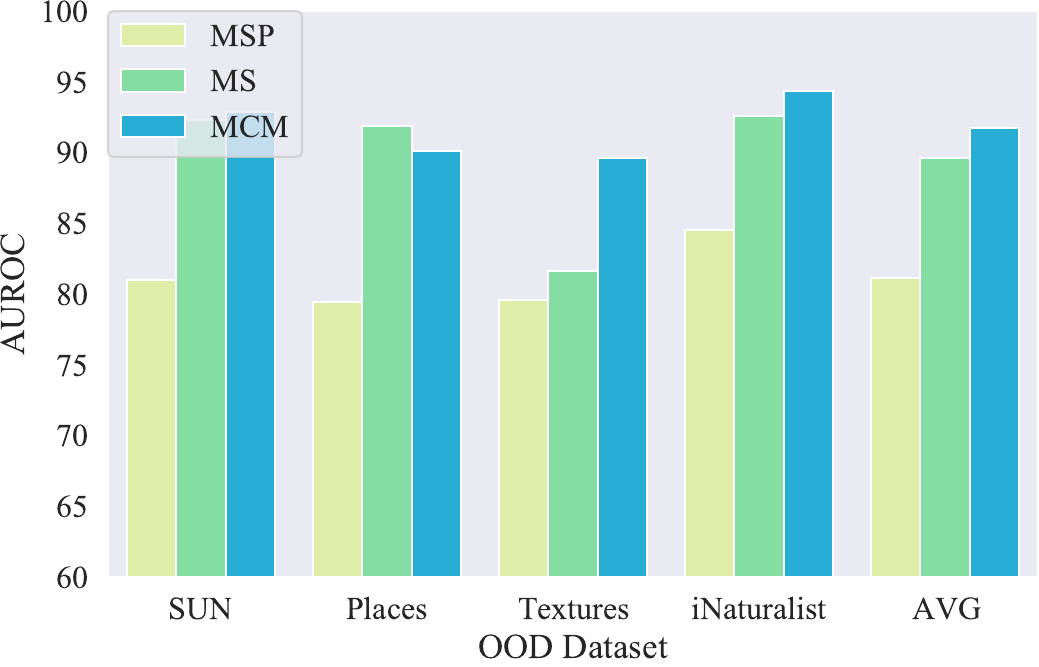}
\caption{\textcolor{cb}{OOD detection performance (AUROC) on ImageNet-1k (ID). The trend is consistent with Fig~\ref{fig:imagenet_fpr}.} }
\label{fig:imagenet_auroc}
\end{figure}

\noindent\textbf{The impact of fine-tuning on feature geometry.}
To better understand how fine-tuning leads to improved OOD detection performance, we examine the geometry of the feature representations. \textcolor{cb}{For illustration, we use the simple $S_{\text{MS}}$ score as it provides an intuitive geometric interpretation}. For each test input, $S_{\text{MS}}$ captures the angular distance between its visual features and the closest ID prototype. Figure~\ref{fig:id_ood_sep} shows $S_{\text{MS}}$ for ID and each OOD test dataset, where radians are converted to degrees for better readability. Intuitively, one desires to learn compact ID clusters such that ID inputs are closer to the nearest ID prototypes than OOD inputs. We illustrate the effects of prompt learning in Figure~\ref{fig:illustrate_sep}. Compared to zero-shot CLIP, \texttt{CoOp} and \texttt{CoCoOp} decrease the angular distance for ID inputs to the nearest concept prototype while simultaneously increasing the angular distance for OOD inputs.  In particular, \texttt{CoCoOp} decreases the angular distance for ID inputs more significantly, resulting in better ID-OOD separability.
Although prompt learning methods introduce perturbations to the feature space, the overall effect is modest, with only a slight deviation of a few degrees from the pre-trained model\footnote{\textcolor{cb}{Similar observations can also be verified for adaptor-based methods.}}. Nonetheless, these perturbations play a crucial role in enhancing both ID classification and OOD detection performance.

\vspace{0.2cm}
\noindent\textbf{\textcolor{cb}{Exploring prompt learning for OOD detection on challenging large-scale benchmarks}} 
In previous sections, we show that prompt learning with both $S_{\text{MS}}$ and $S_{\text{MCM}}$ scores display competitive performance. Next, we consider a more challenging large-scale benchmark ImageNet-1k (ID). The results \textcolor{cb}{in FPR95 and AUROC} are shown in Figure~\ref{fig:imagenet_fpr} and Figure~\ref{fig:imagenet_auroc}. While $S_{\text{MS}}$ outperforms $S_{\text{MSP}}$ score, we can clearly see that $S_{\text{MCM}}$ is particularly advantageous compared to the simpler $S_{\text{MS}}$ baseline. In particular, $S_{\text{MCM}}$ outperforms $S_{\text{MS}}$ by 7.44\% in FPR95 averaged across the four OOD test sets. 
Moreover, \texttt{CoOp} with $S_{\text{MCM}}$ achieves an average FPR95 of 37.74\% on the benchmark, surpassing the zero-shot performance of the large backbone CLIP-L/14 model which has an FPR95 of 38.17\%~\citep{ming2022delving}. These results further demonstrate the effectiveness of $S_{\text{MCM}}$ in CLIP-based prompt learning for challenging scenarios.

\begin{figure}[ht]
  \centering
    \includegraphics[width=0.95\linewidth]{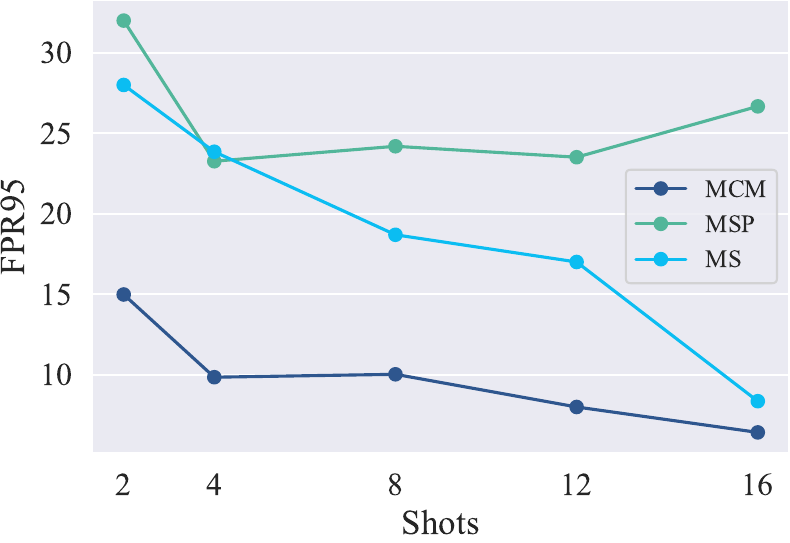}
\caption{The effects of shots for \texttt{CoOp} with various OOD detection scores on Caltech-101 (ID). The performance is averaged over the four OOD test sets.}
\label{fig:shots_coop}
\end{figure}

\begin{figure}[ht]
  \centering
    \includegraphics[width=0.95\linewidth]{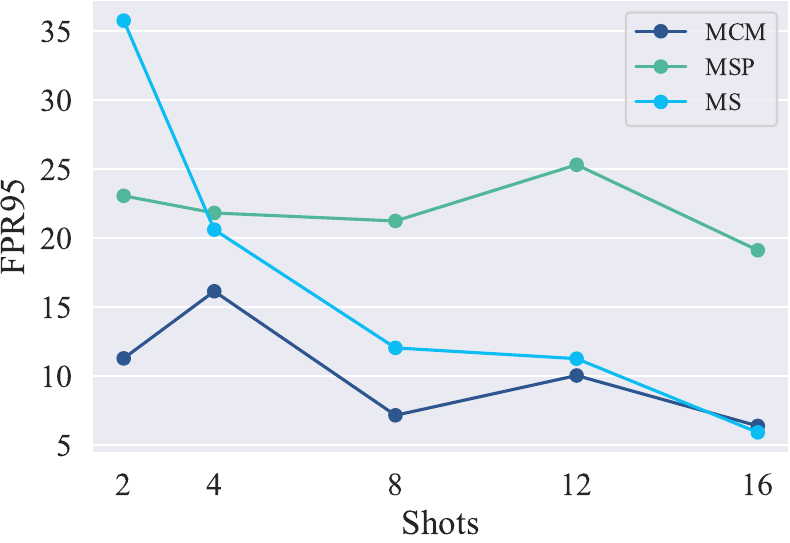}
\caption{The effects of shots for \texttt{CoCoOp} with various OOD detection scores on Caltech-101 (ID). The performance is averaged over the four OOD test sets.}
\label{fig:shots_cocoop}
\end{figure}

\vspace{0.2cm}
\noindent\textbf{The impact of shots.} We investigate the impact of shots for \texttt{CoOp} and \texttt{CoCoOp} with various OOD detection scores. The results are shown in Figure~\ref{fig:shots_coop} and Figure~\ref{fig:shots_cocoop}, where each point represents the average FPR95 over the four OOD test sets. We highlight two key findings. First, the OOD detection performance with both $S_{\text{MS}}$ and $S_{\text{MCM}}$ score improves as the number of shots increases. This trend is consistent with the ID classification accuracy reported in~\cite{zhou2022coop}, suggesting that using a suitable OOD uncertainty score can enhance the representation quality as more data is incorporated during prompt learning. Second, the performance of $S_{\text{MCM}}$ is promising even with a low number of shots, demonstrating its effectiveness in resource-constrained settings.

\begin{table}[!htb]
\caption{The impact of model architecture on ResNet backbones with CoOp on Caltech-101 (ID).}
\label{tab:arch_r}
\centering
\resizebox{0.5\textwidth}{!}{
\begin{tabular}{lllcc}
\toprule
\textbf{Arch}                    & \textbf{Score}                & \textbf{OOD Dataset} & \textbf{FPR95$\downarrow$}         & \textbf{AUROC$\uparrow$} \\ \midrule
\multirow{15}{*}{RN50}      & \multirow{5}{*}{$S_{\text{MSP}}$} & SUN         & 29.93 & 93.95 \\
                            &                      & Places      & 37.64 & 91.96 \\
                            &                      &  Textures         & 35.69 & 93.58 \\
                            &                      & iNaturalist & 43.42 & 91.27 \\
                            &                      &\cellcolor{gray!20} AVG         & \cellcolor{gray!20}36.67 &\cellcolor{gray!20} 92.69 \\ \cmidrule(lr){3-5} 
                            & \multirow{5}{*}{$S_{\text{MS}}$}  & SUN         & 6.02  & 98.45 \\
                            &                      & Places      & 9.02  & 97.79 \\
                            &                      &  Textures         & 23.17 & 95.25 \\
                            &                      & iNaturalist & 12.39 & 97.37 \\
                            &                      & \cellcolor{gray!20}AVG         & \cellcolor{gray!20}12.65 &\cellcolor{gray!20} 97.22 \\\cmidrule(lr){3-5} 
                            & \multirow{5}{*}{$S_{\text{MCM}}$} & SUN         & 8.56  & 98.03 \\
                            &                      & Places      & 17.02 & 95.88 \\
                            &                      &  Textures         & 12.09 & 97.56 \\
                            &                      & iNaturalist & 21.00 & 95.93 \\
                            &                      & \cellcolor{gray!20}AVG         & \cellcolor{gray!20}14.67 &\cellcolor{gray!20} 96.85 \\ \midrule
\multirow{15}{*}{RN101}     & \multirow{5}{*}{$S_{\text{MSP}}$} & SUN         & 23.60 & 95.20 \\
                            &                      & Places      & 29.37 & 93.94 \\
                            &                      &  Textures        & 21.29 & 96.24 \\
                            &                      & iNaturalist & 34.18 & 94.05 \\
                            &                      & \cellcolor{gray!20}AVG         &\cellcolor{gray!20} 27.11 &\cellcolor{gray!20} 94.86 \\ \cmidrule(lr){3-5}
                            & \multirow{5}{*}{$S_{\text{MS}}$}  & SUN         & 19.08 & 96.56 \\
                            &                      & Places      & 20.79 & 96.25 \\
                            &                      &  Textures         & 36.97 & 94.39 \\
                            &                      & iNaturalist & 30.89 & 95.41 \\
                            &                      & \cellcolor{gray!20}AVG         & \cellcolor{gray!20}26.93 & \cellcolor{gray!20}95.65 \\ \cmidrule(lr){3-5}
                            & \multirow{5}{*}{$S_{\text{MCM}}$} & SUN         & 6.19  & 98.42 \\
                            &                      & Places      & 11.57 & 97.16 \\
                            &                      &  Textures         & 5.83  & 98.49 \\
                            &                      & iNaturalist & 10.56 & 97.69 \\
                            &                      &\cellcolor{gray!20} AVG         &\cellcolor{gray!20} 8.54  &\cellcolor{gray!20} 97.94 \\ 
                            \bottomrule
\end{tabular}
}
\end{table}

\begin{table}[!htb]
\caption{The impact of model architecture on ViT backbones with CoOp on Caltech-101 (ID).}
\label{tab:arch_v}
\centering
\resizebox{0.5\textwidth}{!}{
\begin{tabular}{lllcc}
\toprule
\textbf{Arch}                    & \textbf{Score}                & \textbf{OOD Dataset} & \textbf{FPR95$\downarrow$}         & \textbf{AUROC$\uparrow$} \\ \midrule
\multirow{15}{*}{CLIP-B/32} & \multirow{5}{*}{$S_{\text{MSP}}$} & SUN         & 24.20 & 96.02 \\
                            &                      & Places      & 27.94 & 94.99 \\
                            &                      & Textures         & 24.54 & 96.09 \\
                            &                      & iNaturalist & 28.90 & 95.37 \\
                            &                      & \cellcolor{gray!20}AVG         & \cellcolor{gray!20}26.40 &\cellcolor{gray!20} 95.62 \\ \cmidrule(lr){3-5}
                            & \multirow{5}{*}{$S_{\text{MS}}$}  & SUN         & 13.81 & 97.41 \\
                            &                      & Places      & 16.49 & 96.48 \\
                            &                      &  Textures         & 25.23 & 95.24 \\
                            &                      & iNaturalist & 13.00 & 97.60 \\
                            &                      & \cellcolor{gray!20}AVG         & \cellcolor{gray!20}17.13 &\cellcolor{gray!20} 96.68 \\
                            \cmidrule(lr){3-5}
                            & \multirow{5}{*}{$S_{\text{MCM}}$} 
                            & SUN         & 4.06  & 98.92 \\
                            &                      & Places      & 7.31  & 98.01 \\
                            &                      &  Textures         & 4.61  & 98.81 \\
                            &                      & iNaturalist & 8.70  & 98.17 \\
                            &                      & \cellcolor{gray!20}AVG         & \cellcolor{gray!20}6.17  &\cellcolor{gray!20} 98.48 \\
                            \midrule
                            \multirow{15}{*}{CLIP-L/14} & \multirow{5}{*}{$S_{\text{MSP}}$}   & SUN         & 7.73  & 98.36 \\
    && Places      & 10.96 & 97.71 \\
    && Textures         & 19.18 & 96.60 \\
    && iNaturalist & 11.33 & 97.71 \\
                            &                      & \cellcolor{gray!20}AVG         & \cellcolor{gray!20}15.85 &\cellcolor{gray!20} 97.41 \\ \cmidrule(lr){3-5}
                            & \multirow{5}{*}{$S_{\text{MS}}$}  
                            & SUN         & 13.81 & 97.41 \\
                            &                      & Places      & 16.49 & 96.48 \\
                            &                      &  Textures         & 25.23 & 95.24 \\
                            &                      & iNaturalist & 13.00 & 97.60 \\
                            &                      & \cellcolor{gray!20}AVG         & \cellcolor{gray!20} 12.30 &\cellcolor{gray!20} 97.59  \\
                            \cmidrule(lr){3-5}
                            & \multirow{5}{*}{$S_{\text{MCM}}$} 
                             & SUN         & 2.15  & 99.33 \\
    && Places      & 5.60  & 98.30 \\
    && Textures        & 2.32  & 99.31 \\
    && iNaturalist & 3.94  & 99.06 \\
                            &                      & \cellcolor{gray!20}AVG         & \cellcolor{gray!20}3.50   &\cellcolor{gray!20} 99.00 \\
                            \bottomrule
\end{tabular}
}
\end{table}

\vspace{0.2cm}
\noindent\textbf{The impact of backbone architecture.} We conduct another ablation study on the impact of model architectures. We consider CLIP with ResNet backbones (N50, RN101) and \textcolor{cb}{ViT backbones (CLIP-B/32, CLIP-L/14), where the vision encoder is based on ViT-B/32 and ViT-L/14, respectively}. We train with \texttt{CoOp} with hyperparameters following the original implementation for each architecture~\citep{zhou2022coop}. We evaluate the models using $S_{\text{MSP}}$, $S_{\text{MS}}$, and $S_{\text{MCM}}$ score and summarize the results in Table~\ref{tab:arch_r} and Table~\ref{tab:arch_v}. Interestingly, compared to $S_{\text{MSP}}$, $S_{\text{MS}}$ brings more significant improvements under ViT backbones than ResNet backbones. 
In contrast, $S_{\text{MCM}}$ score consistently demonstrates competitive performance for all the architectures considered. For instance, with CLIP-B/32, $S_{\text{MCM}}$ achieves an average FPR95 of 6.17\%, a 20.23\% improvement over the $S_{\text{MSP}}$ baseline. We observe similar improvements for RN101 (18.57\%) and RN50 (22\%). \textcolor{cb}{Moreover, larger backbones lead to superior performance when fixing the OOD detection score as MCM. For example, with CLIP-L/14, the average FPR95 is improved by 11.17\% compared to RN50 and 2.67\% compared to CLIP-B/32. A similar trend has been shown for ID classification~\citep{radford2021learning}, where larger models yield better feature representation.}

\section{Related works}\label{related_works}

\noindent\textbf{Parameter-efficient fine-tuning of vision-language models.} Large-scale vision-language models have shown impressive performance on various downstream tasks~\citep{radford2021learning,jia2021scaling,yao2021filip,li2021supervision}. These models learn transferable feature representations via pre-training on web-scale heterogeneous datasets. However, as downstream datasets can have a limited number of samples, adapting these large models in a parameter and data-efficient manner is crucial for effective knowledge transfer. Recent works propose various ways to tackle this challenge. ~\cite{zhou2022coop} propose to tune a set of soft prompts~\citep{li-liang-2021-prefix, lester-etal-2021-power} while freezing the encoders of CLIP. ~\cite{zhou2022cocoop} aims to improve the generalization ability of CoOp by introducing a meta-network that learns input-dependent tokens. ~\cite{huang2022unsupervised} propose to learn prompts in an unsupervised manner while TPT~\citep{shu2022tpt} uses test-time prompt tuning to learn adaptive prompts on the fly. Beyond textual prompt learning, \cite{bahng2022visual} propose to tune visual prompts for CLIP-based fine-tuning. Another line of work focuses on adaptor-style fine-tuning, where instead of tuning prompts, the feature embedding is directly optimized using an adaptor module~\citep{gao2021clip,zhang2022tip,udandarao2022sus}.  Prior works demonstrate significant improvement over zero-shot CLIP for few-shot ID classification and OOD generalization where OOD labels are given. However, it is unclear how reliable these parameter-efficient fine-tuning methods are for OOD detection tasks. Our work bridges this gap and explores how fine-tuning impacts OOD detection for few-shot downstream datasets.

\vspace{0.2cm}
\noindent\textbf{OOD detection with vision-language representations.}
A plethora of OOD detection methods have been proposed on visual inputs~\citep{lee2018simple,liang2018enhancing,dan2019self,tack2020csi, sun2022knn,ming2022posterior,du2022siren,wang2022vim,ming2023cider}. With the rise of large-scale pre-trained models on vision language inputs, an increasing number of works utilize textual information for visual OOD detection and demonstrate promising performance. \cite{fort2021exploring}  propose a scheme where pre-trained CLIP models are provided with candidate OOD labels for each target dataset, and show that the output probabilities summed over the OOD labels effectively capture OOD uncertainty. 
Without the assumption of OOD labels, \cite{esmaeilpour2022zero} propose to train a decoder based on the visual encoder of CLIP to generate candidate labels for OOD detection. However, training a high-quality decoder incurs significant computational costs and requires extra data. While both \cite{esmaeilpour2022zero} and \cite{radford2021learning} focus on small-scale inputs, \cite{ming2022delving} propose an OOD label-free method MCM which demonstrates promising results on a wide range of large-scale and challenging tasks~\citep{ming2022impact}. However, \cite{ming2022delving} only investigate pre-trained CLIP models. For multi-modal OOD detection benchmarks,~\cite{bitterwolf2023ninco} curate a new OOD test set for ImageNet-1k  while~\cite{gu2023critical} provide new OOD datasets for document understanding.    
In contrast, our work focuses on the impact of parameter-efficient fine-tuning methods for OOD detection in few-shot downstream tasks, which has not been explored.

\section{Conclusion}\label{conclusion}
In this paper, we provide a timely study on the impact of parameter-efficient fine-tuning methods for OOD detection with large vision-language models. We focus on the few-shot setting without access to OOD labels, which has been largely unexplored in the literature. We show that parameter-efficient fine-tuning methods can improve both ID and OOD performance when coupled with a proper OOD score, with prompt learning-based methods showing the strongest performance under the MCM (PEFT-MCM) score. We analyze the feature space and provide insights into the effectiveness of such methods through the lens of multi-modal concept matching. We hope our findings will inspire and motivate future research on designing reliable fine-tuning methods for large vision-language models.

\section*{Acknowledgement}
Support for this
research was provided by American Family Insurance through a research partnership with the
University of Wisconsin–Madison’s Data Science Institute, Office of Naval Research under award number N00014-23-1-2643, AFOSR Young Investigator Program under award number FA9550-23-1-0184, and National Science Foundation (NSF) CAREER Award No. IIS-2237037.

\begin{appendices}

 \section{Dataset Details}\label{secA1}
\textbf{Details on ID and OOD dataset construction} For ID datasets, we follow the same construction as in previous works~\citep{zhang2022tip,zhou2022cocoop,zhou2022coop}. Detailed instructions on dataset installation can be found in \url{https://github.com/KaiyangZhou/CoOp/blob/main/DATASETS.md}. For OOD datasets,~\cite{huang2021mos} curate a collection of subsets from iNaturalist~\cite{van2018inaturalist}, SUN~\cite{xiao2010sun}, Places~\cite{zhou2017places}, and Texture~\cite{cimpoi2014describing} as large-scale OOD datasets for ImageNet-1k, where the classes of the test sets do not overlap with ImageNet-1k. Detailed instructions can be found in \url{https://github.com/deeplearning-wisc/large_scale_ood}.

\section{Additional Results}\label{secA2}
\subsection{ID accuracy} While we primarily focus on the OOD detection performance of CLIP-based fine-tuning methods, we present the results of the ID accuracy for each dataset based on CLIP-B/16 in Table~\ref{tab:id_acc} for completeness. Further results on the ID accuracy with various datasets and architectures can be seen in~\cite{zhou2022cocoop},~\cite{zhou2022coop}, and ~\cite{zhang2022tip}.

\begin{table}[!htb]
\caption{ID accuracy on the downstream datasets for CLIP-based fine-tuning methods with CLIP-B/16.}
\label{tab:id_acc}
\centering
\resizebox{0.4\textwidth}{!}{
\begin{tabular}{llc}
\toprule
\textbf{ID Dataset}                     & \textbf{Method}      & \textbf{ID Acc} \\
\midrule
\multirow{5}{*}{Caltech-101}   & ZOCLIP      & 92.90  \\
                               & TipAdaptor  & 95.01  \\
                               & TipAdaptorF & 95.66  \\
                               & CoOp        & 95.30  \\
                               & CoCoOp      & 95.00  \\
                               \midrule
\multirow{5}{*}{Food-101}      & ZOCLIP      & 86.10  \\
                               & TipAdaptor  & 86.49  \\
                               & TipAdaptorF & 87.43  \\
                               & CoOp        & 85.50  \\
                               & CoCoOp      & 87.30  \\
                               \midrule
\multirow{5}{*}{Stanford-Cars} & ZOCLIP      & 65.27  \\
                               & TipAdaptor  & 75.29  \\
                               & TipAdaptorF & 83.40  \\
                               & CoOp        & 78.50  \\
                               & CoCoOp      & 72.30  \\
                               \midrule
\multirow{5}{*}{Oxford-Pets}   & ZOCLIP      & 89.10  \\
                               & TipAdaptor  & 91.85  \\
                               & TipAdaptorF & 92.91  \\
                               & CoOp        & 93.40  \\
                               & CoCoOp      & 93.30  \\
                               \midrule
\multirow{5}{*}{ImageNet-1k}   & ZOCLIP      & 68.77  \\
                               & TipAdaptor  & 70.26  \\
                               & TipAdaptorF & 73.70  \\
                               & CoOp        & 71.63  \\
                               & CoCoOp      & 71.20  \\
                               \bottomrule
\end{tabular}
}
\end{table}

\subsection{OOD detection performance based on visual features alone}
In this section, we explore several commonly used OOD detection scores solely based on the visual branch of CLIP models. Specifically, we consider the Mahalanobis score~\citep{lee2018simple} 
on the penultimate layer of the visual encoder and MSP~\citep{hendrycks2016baseline}, Energy~\citep{liu2020energy}, and KL Matching~\citep{hendrycks2022scaling} scores on the logit layer after linear probing the visual encoder. The results are summarized in Table~\ref{tab:visual-only}, based on 16-shot Caltech-101 (ID). We can see that the Mahalanobis score does not yield promising performance because 1) the feature embeddings from the visual encoder of CLIP may not follow class-conditional Gaussian distributions, 2) it is challenging to estimate the mean and especially covariance matrix when the number of samples is much smaller than the feature dimension in the few-shot setting. On the other hand, the OOD scores based on fine-tuned logit layer result in worse performance compared to the MCM score. One major reason is that fine-tuning CLIP in the few-shot setting is prone to overfitting the downstream ID dataset, making the model less reliable. This further highlights the importance of choosing OOD detection scores fitted to parameter-efficient fine-tuning methods.

\begin{table}[t]
\caption{Additional results for OOD scores based on visual encoder only. ID dataset is Caltech-101 (16 shot).}
\label{tab:visual-only}
\centering
\resizebox{0.48\textwidth}{!}{
\begin{tabular}{llcc}
\toprule
\textbf{OOD Score}                & \textbf{OOD Dataset} & \textbf{FPR95$\downarrow$}         & \textbf{AUROC$\uparrow$} \\ \midrule
\multirow{5}{*}{Maha}        & SUN         & 34.15 & 95.20  \\
                             & Places      & 20.50  & 96.21 \\
                             & Textures    & 64.10  & 92.43 \\
                             & iNaturalist & 66.62 & 92.97 \\
                             & \cellcolor{gray!20}AVG         & \cellcolor{gray!20}46.34 & \cellcolor{gray!20}94.20  \\
                             \midrule
\multirow{5}{*}{Energy}      & SUN         & 15.02 & 97.05 \\
                             & Places      & 21.10  & 95.75 \\
                             & Textures    & 15.60  & 97.00  \\
                             & iNaturalist & 33.77 & 95.49 \\
                             & \cellcolor{gray!20}AVG         & \cellcolor{gray!20}21.37 &\cellcolor{gray!20} 96.32 \\
                             \midrule
\multirow{5}{*}{KL Matching} & SUN         & 4.56  & 98.21 \\
                             & Places      & 8.92  & 97.52 \\
                             & Textures    & 42.64 & 94.47 \\
                             & iNaturalist & 9.70   & 97.35 \\
                             & \cellcolor{gray!20}AVG         & \cellcolor{gray!20}16.46 &\cellcolor{gray!20} 96.89 \\
                             \midrule
\multirow{5}{*}{MSP}         & SUN         & 16.23 & 96.59 \\
                             & Places      & 20.98 & 95.97 \\
                             & Textures    & 7.15  & 98.33 \\
                             & iNaturalist & 11.79 & 97.31 \\
                             & \cellcolor{gray!20}AVG         & \cellcolor{gray!20}14.04 & \cellcolor{gray!20}97.05\\
                             \bottomrule
\end{tabular}
}
\end{table}

\subsection{Additional results on ImageNet-1k}\label{sec:1k}
In this section, we consider two additional OOD test sets ImageNet-O~\citep{hendrycks2021nae} and OpenImage-O~\citep{wang2022vim} for ImageNet-1k (ID). OpenImage-O is a subset curated from the test set of OpenImage-V3~\citep{openimages} containing a diverse set of categories. ImageNet-O is a challenging OOD dataset that contains naturally adversarial examples for ImageNet-1k. 
The results are shown in Table~\ref{tab:1k-add}. The model (CLIP-B/16) is trained with CoOp. We can see that: 1) The performance on ImageNet-O is generally worse than the rest of OOD test sets (iNaturalist, Textures, SUN, Places) in Section~\ref{sec:delve}, suggesting that this task remains challenging in the context of few-shot prompt learning. 2) MCM score still performs the best compared to MS and MSP on both OOD test sets, consistent with our previous observations, which further highlights the importance of softmax and temperature scaling for OOD detection with fine-tuning.

\begin{table}[!htb]
\caption{OOD detection performance on two OOD additional test sets for ImageNet-1k (ID). We train CLIP-B/16 with CoOp.}
\label{tab:1k-add}
\centering
\resizebox{0.47\textwidth}{!}{
\begin{tabular}{lccc}
\toprule
\textbf{OOD Dataset} & \textbf{OOD Score} & \textbf{FPR95$\downarrow$}         & \textbf{AUROC$\uparrow$} \\ \midrule
\multirow{3}{*}{ImageNet-O}  
                              & $S_{\text{MSP}}$       & 77.20  & 74.01 \\
                              & $S_{\text{MS}}$        & 70.75 & 82.30  \\
                               & $S_{\text{MCM}}$       & 61.50  & 84.13 \\ \midrule
\multirow{3}{*}{OpenImage-O} 
                              & $S_{\text{MSP}}$       & 56.89 & 83.73 \\
                              & $S_{\text{MS}}$        & 39.18 & 91.48 \\
                              & $S_{\text{MCM}}$       & 36.68 & 92.76 \\
                              \bottomrule
\end{tabular}
}
\end{table}

\vspace{-0.7cm}\subsection{Alternative OOD scores}\label{sec:add_score}
In this section, we investigate the performance with several alternative OOD scoring functions based on the cosine similarities of input $\*x$ with the $k$-th label $s_k(\*x)$, $k\in \{1,2,...,K\}$ (defined in Section~\ref{sec:score}). Specifically, we consider the energy and the KL matching score for each adaptation method and summarize the results based on Caltech-101 (ID) in Table~\ref{tab:add_comparison}.  We observe that 1) using the energy score, all adaptation methods significantly enhance the performance over the zero-shot baseline (ZOCLIP). 2) the general performance vastly improves when utilizing the KL Matching score. However, even the highest achieved performance (FPR95 at 7.91 with CoCoOp) falls short when compared to the MCM score (FPR95 at 5.02 with CoCoOp).

\begin{table*}[!htb]
\caption{OOD detection performance based on $S_{\text{MSP}}$ score. The average performance for most adaptation methods is much worse than using $S_{\text{MS}}$ (Table~\ref{tab:fine-grain}) and $S_{\text{MCM}}$ (Table~\ref{tab:fine-grain_mcm}).}
\label{tab:msp}
\centering
\resizebox{0.95\textwidth}{!}{
\begin{tabular}{llcccccccccc}
\toprule
\multirow{2}{*}{\textbf{ID Dataset}}           & \multirow{2}{*}{\textbf{Method}} & \multicolumn{2}{c}{\textbf{SUN}} & \multicolumn{2}{c}{\textbf{Places}} & \multicolumn{2}{c}{\textbf{Textures}} & \multicolumn{2}{c}{\textbf{iNaturalist}} & \multicolumn{2}{c}{\textbf{Average}} \\
\cmidrule(lr){3-4} \cmidrule(lr){5-6} \cmidrule(lr){7-8} \cmidrule(lr){9-10} \cmidrule(lr){11-12}
                              &                         & \textbf{FPR95$\downarrow$}         & \textbf{AUROC$\uparrow$}     & \textbf{FPR95$\downarrow$}         & \textbf{AUROC$\uparrow$}       & \textbf{FPR95$\downarrow$}         & \textbf{AUROC$\uparrow$}         & \textbf{FPR95$\downarrow$}         & \textbf{AUROC$\uparrow$}          & \textbf{FPR95$\downarrow$}         & \textbf{AUROC$\uparrow$}      \\
                              \midrule
                              &                         & FPR95      & AUROC      & FPR95        & AUROC       & FPR95         & AUROC        & FPR95          & AUROC          & FPR95      & AUROC      \\
                              \midrule
\multirow{5}{*}{Food-101}      & ZOCLIP                  & 11.48      & 97.76      & 13.11        & 97.48       & 15.04         & 96.08        & 16.65          & 96.73          & 14.07      & 97.01      \\
                              & TipAdaptor              & 7.32       & 98.51      & 9.03         & 98.31       & 11.88         & 96.94        & 14.47          & 97.21          & \textbf{10.68}      & \textbf{97.74}      \\
                              & TipAdaptorF             & 15.08      & 97.26      & 15.38        & 97.24       & 17.57         & 95.99        & 20.95          & 96.18          & 17.25      & 96.67      \\
                              & CoOp                    & 19.66      & 96.20      & 21.15        & 95.95       & 28.33         & 93.62        & 23.80          & 95.51          & 23.23      & 95.32      \\
                              & CoCoOp                  & 8.67       & 98.28      & 10.56        & 98.03       & 14.77         & 96.23        & 14.33          & 97.26          & 12.08      & 97.45      \\
                              \midrule
\multirow{5}{*}{Oxford-Pets} & ZOCLIP                  & 24.67      & 94.72      & 28.54        & 93.71       & 19.01         & 96.42        & 39.77          & 93.01          & 28.00      & 94.47      \\
                              & TipAdaptor              & 15.66      & 97.11      & 18.83        & 96.45       & 12.50         & 97.92        & 25.19          & 95.90          & 18.04      & 96.84      \\
                              & TipAdaptorF             & 16.79      & 96.77      & 20.33        & 96.04       & 12.22         & 97.90        & 26.62          & 95.80          & 18.99      & 96.63      \\
                              & CoOp                    & 8.46       & 98.50      & 10.75        & 98.13       & 11.21         & 98.09        & 32.13          & 94.08          & 15.64      & 97.20      \\
                              & CoCoOp                  & 9.06       & 98.31      & 10.43        & 98.13       & 7.39          & 98.70        & 27.97          & 95.11          & \textbf{13.71}      & \textbf{97.56}      \\
                              \midrule
\multirow{5}{*}{Stanford-Cars} & ZOCLIP                  & 6.99       & 98.49      & 10.33        & 97.68       & 8.24          & 98.39        & 32.85          & 92.56          & 14.60      & 96.78      \\
                              & TipAdaptor              & 1.94       & 99.58      & 3.30         & 99.31       & 1.97          & 99.56        & 12.52          & 97.80          & \textbf{4.93}       & \textbf{99.06}      \\
                              & TipAdaptorF             & 15.39      & 97.19      & 14.01        & 97.32       & 8.39          & 98.49        & 21.88          & 95.90          & 14.92      & 97.22      \\
                              & CoOp                    & 9.88       & 98.05      & 14.07        & 97.12       & 10.71         & 97.71        & 36.73          & 91.51          & 17.85      & 96.10      \\
                              & CoCoOp                  & 9.99       & 97.81      & 11.87        & 97.15       & 10.46         & 97.69        & 31.58          & 92.59          & 15.97      & 96.31   \\
                              \midrule
                              \multirow{5}{*}{Caltech-101}   & ZOCLIP                  & 16.17      & 96.47      & 22.45        & 94.96       & 17.89         & 96.33        & 15.01          & 96.96          & 17.88      & 96.18      \\
                              & TipAdaptor              & 12.98      & 97.40      & 17.79        & 96.77       & 13.74         & 97.72        & 20.08          & 96.65          & \textbf{16.15}      & \textbf{97.13}      \\
                              & TipAdaptorF             & 17.94      & 96.68      & 22.92        & 95.74       & 15.16         & 97.40        & 24.18          & 96.01          & 20.05      & 96.46      \\
                              & CoOp                    & 24.07      & 96.11      & 29.91        & 94.59       & 26.29         & 95.72        & 26.35          & 95.92          & 26.66      & 95.58      \\
                              & CoCoOp                  & 14.92      & 97.32      & 20.67        & 95.91       & 19.20         & 96.56        & 21.74          & 96.33          & 19.13      & 96.53      \\
                              
                              \bottomrule
\end{tabular}
}
\end{table*}

\begin{table*}[!htb]
\caption{Comparison with additional OOD scores on Caltech-101 (ID). $S_{\text{KL}}$ stands for the KL matching score~\citep{hendrycks2022scaling} and $S_{\text{Energy}}$ denotes the energy score~\citep{liu2020energy}.
}
\label{tab:add_comparison}
\centering
\resizebox{0.95\textwidth}{!}{
\begin{tabular}{clcccccccccc}
\toprule
\multirow{2}{*}{\textbf{OOD Score}}           & \multirow{2}{*}{\textbf{Method}} & \multicolumn{2}{c}{\textbf{SUN}} & \multicolumn{2}{c}{\textbf{Places}} & \multicolumn{2}{c}{\textbf{Textures}} & \multicolumn{2}{c}{\textbf{iNaturalist}} & \multicolumn{2}{c}{\textbf{Average}} \\
\cmidrule(lr){3-4} \cmidrule(lr){5-6} \cmidrule(lr){7-8} \cmidrule(lr){9-10} \cmidrule(lr){11-12}
                              &                         & \textbf{FPR95$\downarrow$}         & \textbf{AUROC$\uparrow$}     & \textbf{FPR95$\downarrow$}         & \textbf{AUROC$\uparrow$}       & \textbf{FPR95$\downarrow$}         & \textbf{AUROC$\uparrow$}         & \textbf{FPR95$\downarrow$}         & \textbf{AUROC$\uparrow$}          & \textbf{FPR95$\downarrow$}         & \textbf{AUROC$\uparrow$}      \\
                              \midrule
                              \multirow{5}{*}{$S_{\text{MS}}$}                                & ZOCLIP                  & 32.03      & 94.06      & 33.01        & 93.39       & 54.66        & 89.29         & 32.14          & 94.30          & 37.96      & 92.76      \\
                              & TipAdaptor                 & 9.69       & 98.07      & 11.25        & 97.84       & 20.90        & 96.68         & 13.62          & 97.72          & 13.86      & 97.58      \\
                              & TipAdaptorF                & 10.20      & 97.76      & 11.60        & 97.42       & 23.32        & 95.54         & 14.01          & 97.36          & 14.78      & 97.02      \\
                              & CoOp                    & 5.53       & 98.56      & 9.88         & 97.50       & 13.10        & 97.10         & 4.89           & 98.76          & 8.35       & 97.98      \\
                              & CoCoOp                  & 2.86       & 99.19      & 6.42         & 98.37       & 8.81         & 98.09         & 5.68           & 98.68          & 5.94       & 98.58      \\
                               \midrule
                              \multirow{5}{*}{$S_{\text{MCM}}$}                & ZOCLIP                 &14.83&	97.20 &20.45&	96.00 &	14.98&	97.35	&10.84&	97.76&	15.28&	97.08    \\
& TipAdaptor    &  5.12	&98.83    &8.05&	98.34&	4.65&	99.05&	6.94&	98.77&	6.19	&\textbf{98.75}   \\
                              & TipAdaptorF &4.83	&98.79 & 8.09&	98.07	&6.41	&98.11	&4.94	&98.98&	6.07&	98.49                  \\
                
                              & CoOp    &     3.62& 	99.01	& 8.15& 	97.89	& 6.29& 	98.62& 	7.57	& 98.35	& 6.41	& 98.47                   \\
                              & CoCoOp   &   4.26&	98.94	&6.76&	98.00&	4.33	&98.88&	4.71&	98.68&	\textbf{5.02}&	98.62                 \\
                                                  \midrule
                              \multirow{5}{*}{$S_{\text{Energy}}$}                &ZOCLIP      & 53.83 & 90.22 & 50.51 & 90.21 & 74.10 & 83.20 & 56.00 & 90.13 & 58.61 & 88.44 \\
&TipAdaptor  & 11.71 & 97.72 & 12.20 & 97.61 & 30.48 & 95.73 & 16.42 & 97.30 & 17.70 & 97.09 \\
&TipAdaptorF & 11.57 & 97.46 & 11.89 & 97.30 & 29.38 & 94.70 & 16.18 & 96.90 & 17.26 & 96.59 \\
&CoOp        & 6.58  & 98.29 & 11.16 & 97.20 & 18.19 & 96.32 & 5.92  & 98.53 & 10.46 & 97.59 \\
&CoCoOp      & 5.22  & 98.87 & 8.80  & 98.13 & 17.30 & 96.87 & 11.28 & 97.95 & 10.65 & 97.95 \\ 
\midrule
\multirow{5}{*}{$S_{\text{KL}}$} 
&ZOCLIP      & 5.51 & 97.57 & 9.48  & 96.61 & 7.41 & 97.64 & 11.43 & 96.22 & 14.02 & 97.31 \\
&TipAdaptor  & 5.54 & 97.63 & 7.69  & 97.13 & 5.74 & 97.96 & 8.00  & 97.37 & 6.74  & 97.52 \\
&TipAdaptorF & 8.52 & 96.89 & 13.00 & 95.92 & 7.02 & 98.02 & 10.71 & 97.11 & 9.81  & 96.98 \\
&CoOp        & 7.15 & 98.06 & 12.37 & 96.60 & 8.74 & 97.62 & 9.33  & 98.00 & 9.40  & 97.57 \\
&CoCoOp      & 4.07 & 98.95 & 9.61  & 97.59 & 5.30 & 98.77 & 12.67 & 97.57 & 7.91  & 98.22 \\
                              \bottomrule
                              
\end{tabular}
}
\end{table*}

\end{appendices}


\bibliography{sn-bibliography}

\end{document}